\documentclass[journal]{IEEEtran}
\usepackage{cite}

\ifCLASSINFOpdf
\else
\fi
\usepackage{amsmath}
\usepackage{graphicx}

\usepackage[utf8]{inputenc}
\usepackage{multirow}
\usepackage{color}

\usepackage{arydshln}
\usepackage[bottom]{footmisc}
\usepackage{rotating}
\usepackage{pdflscape}

\usepackage{bm}
\usepackage{amssymb, bbm} 
\usepackage{afterpage}

\usepackage[caption=false,font=footnotesize]{subfig}
\usepackage{cprotect}
\usepackage{url}
\usepackage[ruled,noresetcount]{algorithm2e}
\usepackage{adjustbox}

\renewcommand{\vec}[1]{\bm{#1}}
\newcommand{\matr}[1]{\bm{#1}}

\newcommand{\norm}[1]{\lVert#1\rVert}
\newcommand{\card}[1]{|#1|}
\newcommand{\Real}{\ensuremath{\mathbb{R}}}
\newcommand{\indicator}{\ensuremath{\mathbbm{1}}}
\definecolor{darkgreen}{rgb}{0.0,0.5,0.0}

\newcommand{\forced}{\color{black}}
\newcommand{\notforced}{\bf}

\begin{document}
%
\title{Imparting Interpretability to Word Embeddings while Preserving Semantic Structure}
%
%
%

\author{Lütfi~Kerem~Şenel,
        İhsan Utlu,
        Furkan~Şahinuç,
        Haldun M. Ozaktas,
        Aykut~Koç

\thanks{L. K. Şenel was with the ASELSAN Research Center, Ankara 06370, Turkey, with the Electrical and Electronics Engineering, Bilkent University, Ankara 06800, Turkey, and also with the UMRAM, Bilkent University, Ankara 06800, Turkey. He is with the CIS, Ludwig Maximillian University, Munich 80538, Germany (e-mail: lksenel@gmail.com).}
\thanks{İhsan Utlu was with the ASELSAN Research Center, Ankara 06370, Turkey.}

\thanks{ F. Şahinuç is the ASELSAN Research Center and also with the Electrical and Electronics Engineering, Bilkent University, Ankara 06800, Turkey.}
\thanks{H. M. Ozaktas is with the Electrical and Electronics Engineering, Bilkent University, Ankara 06800, Turkey.}
\thanks{A. Koç  is with the Electrical and Electronics Engineering, Bilkent University, Ankara 06800, Turkey, and also with the UMRAM, Bilkent University, Ankara 06800, Turkey.}}

\maketitle

\begin{abstract}
As a ubiquitous method in natural language processing, word embeddings are extensively employed to map semantic properties of words into a dense vector representation. They capture semantic and syntactic relations among words but the vectors corresponding to the words are only meaningful relative to each other. Neither the vector nor its dimensions have any absolute, interpretable meaning. We introduce an additive modification to the objective function of the embedding learning algorithm that encourages the embedding vectors of words that are semantically related to a predefined concept to take larger values along a specified dimension, while leaving the original semantic learning mechanism mostly unaffected. In other words, we align words that are already determined to be related, along predefined concepts. Therefore, we impart interpretability to the word embedding by assigning meaning to its vector dimensions. The predefined concepts are derived from an external lexical resource, which in this paper is chosen as Roget's Thesaurus. We observe that alignment along the chosen concepts is not limited to words in the Thesaurus and extends to other related words as well. We quantify the extent of interpretability and assignment of meaning from our experimental results. Manual human evaluation results have also been presented to further verify that the proposed method increases interpretability. We also demonstrate the preservation of semantic coherence of the resulting vector space by using word-analogy/word-similarity tests and a downstream task. These tests show that the interpretability-imparted word embeddings that are obtained by the proposed framework do not sacrifice performances in common benchmark tests.
\end{abstract}


%
\IEEEpeerreviewmaketitle

\section{Introduction}\label{sec:intro}

Distributed word representations, commonly referred to as \emph{word embeddings} \cite{mikolov13word2vec_a,mikolov13word2vec_b,pennington14glove,bojanowski17fasttext}, serve as elementary building blocks in the course of algorithm design for an expanding range of applications in natural language processing (NLP), including named entity recognition \cite{turian10NER,sienvcnik15NER}, parsing \cite{chen14parsing}, sentiment analysis \cite{socher11sentiment,yu17sentiment}, and word-sense disambiguation \cite{iacobacci16disambiguation}. Although the empirical utility of word embeddings as an unsupervised method for capturing the semantic or syntactic features of a certain word as it is used in a given lexical resource is well-established \cite{goldberg17NLPBook,vine15Clinical,jodhi16SarcasmDetection}, an understanding of what these features mean remains an open problem \cite{chen16InfoGAN,levy14dependency} and as such word embeddings mostly remain a black box. It is desirable to be able to develop insight into this black box and be able to interpret what it means, while retaining the utility of word embeddings as semantically-rich intermediate representations. Other than the intrinsic value of this insight, this would not only allow us to explain and understand how algorithms work \cite{goodman17explanation}, but also set a ground that would facilitate the design of new algorithms in a more deliberate way.

Recent approaches to generating word embeddings \cite{mikolov13word2vec_a,pennington14glove} are rooted linguistically in the field of \emph{distributed semantics} \cite{harris54distributional}, where words are taken to assume meaning mainly by their degree of interaction (or lack thereof) with other words in the lexicon \cite{firth57linguistics,firth57synopsis}. Under this paradigm, dense, continuous vector representations are learned in an unsupervised manner from a large corpus, using the word cooccurrence statistics directly or indirectly, and such an approach is shown to result in vector representations that mathematically capture various semantic and syntactic relations between words \cite{mikolov13word2vec_a,pennington14glove,bojanowski17fasttext}. However, the dense nature of the learned embeddings obfuscate the distinct concepts encoded in the different dimensions, which renders the resulting vectors virtually uninterpretable. The learned embeddings make sense only in relation to each other and their specific dimensions do not carry explicit information that can be interpreted. However, being able to interpret a word embedding would illuminate the semantic concepts implicitly represented along the various dimensions of the embedding, and reveal its hidden semantic structures.

In the literature, researchers tackled the interpretability problem of the word embeddings using different approaches. Several researchers \cite{murphy12nnse,luo15online,fyshe14interpretable} proposed algorithms based on non-negative matrix factorization (NMF) applied to cooccurrence variant matrices. Other researchers suggested to obtain interpretable word vectors from existing uninterpretable word vectors by applying sparse coding \cite{arora18linalg,faruqui15sparse}, by training a sparse auto-encoder to transform the embedding space \cite{subramanian18spine}, by rotating the original embeddings \cite{zobnin17rotations,park17rotated} or by applying transformations based on external semantic datasets \cite{senel18semanticStructure}.

Although the above-mentioned approaches provide better interpretability that is measured using a particular method such as word intrusion test, usually the improved interpretability comes with a cost of performance in the benchmark tests such as word similarity or word analogy. One possible explanation for this performance decrease is that the proposed transformations from the original embedding space distort the underlying semantic structure constructed by the original embedding algorithm. Therefore, it can be claimed that a method that learns dense and interpretable word embeddings without inflicting any damage to the underlying semantic learning mechanism is the key to achieve both high performing and interpretable word embeddings.

Especially after the introduction of the \textit{word2vec} algorithm by \cite{mikolov13word2vec_a,mikolov13word2vec_b}, there has been a growing interest in algorithms that generate improved word representations under some performance metric. Significant effort is spent on appropriately modifying the objective functions of the algorithms in order to incorporate knowledge from external resources, with the purpose of increasing the performance of the resulting word representations \cite{yu14Lexical,miller95wordnet,xu14RC_NET,liu15ordinal,bollegala15joint,jauhar15ontologically,johansson15Embedding}. 
Significant effort is also spent on developing retrofitting objectives for the same purpose, independent of the original objectives of the embedding model, to fine-tune the embeddings without joint optimization~\cite{faruqui15retrofitting,mrksic16counterFitting,mrksic17specialization}.
Inspired by the line of work reported in these studies, we propose to use modified objective functions for a different purpose: learning more interpretable dense word embeddings. By doing this, we aim to incorporate semantic information from an external lexical resource into the word embedding so that the embedding dimensions are {\em aligned\/} along predefined concepts. This alignment is achieved by introducing a modification to the embedding learning process. In our proposed method, which is built on top of the GloVe algorithm \cite{pennington14glove}, the cost function for any one of the words of concept word-groups is modified by the introduction of an additive term to the cost function. Each embedding vector dimension is first associated with a concept. For a word belonging to any one of the word-groups representing these concepts, the modified cost term favors an increase for the value of this word's embedding vector dimension corresponding to the concept that the particular word belongs to. For words that do not belong to any one of the word-groups, the cost term is left untouched. Specifically, \emph{Roget's Thesaurus} \cite{Roget1911thesaurus, Roget2008thesaurus} is used to derive the concepts and concept word-groups to be used as the external lexical resource for our proposed method. We quantitatively demonstrate the increase in interpretability by using the measure given in \cite{senel18semanticStructure,senel18trInterpret} as well as demonstrating qualitative results. Furthermore, manual human evaluations based on the ``word intrusion" test given in \cite{chang09wordintrusion} have been carried out for verification. We also show that the semantic structure of the original embedding has not been harmed in the process since there is no performance loss with standard word-similarity or word-analogy tests and with a downstream sentiment analysis task.

The paper is organized as follows. In Section \ref{sec:related}, we discuss previous studies related to our work under two main categories: interpretability of word embeddings and joint-learning frameworks where the objective function is modified. In Section~\ref{sec:problem_description}, we present the problem framework and provide the formulation within the GloVe \cite{pennington14glove} algorithm setting. In Section~\ref{sec:infl} where our approach is proposed, we motivate and develop a modification to the original objective function with the aim of increasing representation interpretability. In Section~\ref{sec:exp}, experimental results are provided and the proposed method is quantitatively and qualitatively evaluated. Additionally, in Section~\ref{sec:exp}, results demonstrating the extent to which the original semantic structure of the embedding space is affected are presented by using word-analogy/word-similarity tests and a downstream evaluation task. Analysis of several parameters of our proposed approach are also presented in Section~\ref{sec:exp}. We conclude the paper in Section~\ref{sec:conclusion}.

\section{Related Work} \label{sec:related}

Methodologically, our work is related to prior studies that aim to obtain ``improved'' word embeddings using external lexical resources, under some performance metric. Previous work in this area can be divided into two main categories: works that i) \emph{modify} the word embedding learning algorithm to incorporate lexical information, ii) operate on pre-trained embeddings with a \emph{post-processing} step.

Among works that follow the first approach, \cite{yu14Lexical} extend the Skip-Gram model by incorporating the word similarity relations extracted from the Paraphrase Database (PPDB) and WordNet~\cite{miller95wordnet}, into the Skip-Gram predictive model as an additional cost term. In \cite{xu14RC_NET}, the authors extend the CBOW model by considering two types of semantic information, termed \emph{relational\/} and \emph{categorical\/}, to be incorporated into the embeddings during training. For the former type of semantic information, the authors propose the learning of explicit vectors for the different relations extracted from a semantic lexicon such that the word pairs that satisfy the same relation are distributed more homogeneously. For the latter, the authors modify the learning objective such that some weighted average distance is minimized for words under the same semantic category. In \cite{liu15ordinal}, the authors represent the synonymy and hypernymy-hyponymy relations in terms of \emph{inequality constraints}, where the pairwise similarity rankings over word triplets are forced to follow an order extracted from a lexical resource. Following their extraction from WordNet, the authors impose these constraints in the form of an additive cost term to the Skip-Gram formulation. Finally, \cite{bollegala15joint} builds on top of the GloVe algorithm by introducing a regularization term to the objective function that encourages the vector representations of similar words as dictated by WordNet to be similar as well.

Turning our attention to the post-processing approach for enriching word embeddings with external lexical knowledge, \cite{faruqui15retrofitting} has introduced the \emph{retrofitting\/} algorithm that acts on pre-trained embeddings such as Skip-Gram or GloVe. The authors propose an objective function that aims to balance out the semantic information captured in the pre-trained embeddings with the constraints derived from lexical resources such as WordNet, PPDB and FrameNet. One of the models proposed in \cite{jauhar15ontologically} extends the retrofitting approach to incorporate the word sense information from WordNet. Similarly, \cite{johansson15Embedding} creates multi-sense embeddings by gathering the word sense information from a lexical resource and learning to decompose the pre-trained embeddings into a convex combination of sense embeddings. In \cite{mrksic16counterFitting}, the authors focus on improving word embeddings for capturing word similarity, as opposed to mere relatedness. To this end, they introduce the \emph{counter-fitting\/} technique which acts on the input word vectors such that synonymous words are attracted to one another whereas antonymous words are repelled, where the synonymy-antonymy relations are extracted from a lexical resource.  The \emph{ATTRACT-REPEL} algorithm proposed by \cite{mrksic17specialization} improves on counter-fitting by a formulation which imparts the word vectors with external lexical information in \emph{mini-batches}. More recently, several global specialization methods have been proposed in order to generalize the specialization to the vectors of words that are not present in external lexical resources \cite{glavas18explicitRetrofitting,ponti18adverserialPropagation}.

Most of the studies discussed above \cite{xu14RC_NET,liu15ordinal,bollegala15joint,faruqui15retrofitting,jauhar15ontologically,mrksic16counterFitting,mrksic17specialization} report performance improvements in benchmark tests such as word similarity or word analogy, while \cite{miller95wordnet} uses a different analysis method (mean reciprocal rank). In sum, the literature is rich with studies aiming to obtain word embeddings that perform better under specific performance metrics. However, less attention has been directed to the issue of interpretability of the word embeddings. In the literature, the problem of interpretability has been tackled using different approaches. In terms of methodology, these approaches can be grouped under two categories: direct approaches that do not require a pre-trained embedding space and post-processing approaches that operate on a pre-trained embedding space (the latter being more often deployed). Among the approaches that fall into the direct approach category, \cite{murphy12nnse} proposed non-negative matrix factorization (NMF) for learning sparse, interpretable word vectors from co-occurrence variant matrices where the resulting vector space is called non-negative sparse embeddings (NNSE). However, since NMF methods require maintaining a global matrix for learning, they suffer from memory and scale issue. This problem has been addressed in \cite{luo15online} where an online method of learning interpretable word embeddings from corpora is proposed. The authors proposed a modified version of skip-gram model \cite{mikolov13word2vec_a}, called OIWE-IPG, where the updates are forced to be non-negative during the training of the algorithm so that the resulting embeddings are also non-negative and more interpretable. In \cite{fyshe14interpretable} a generalized version of NNSE method, called JNNSE, is proposed to incorporate constraints based on external knowledge. In their study, brain activity based word similarity information is taken as external knowledge and combined with text-based similarity information in order to improve interpretability.
 
Relatively more research effort has been directed to improve interpretability by post-processing the existing pre-trained word embeddings. These approaches aim to learn a transformation to map the original embedding space to a new more interpretable one. \cite{arora18linalg} and \cite{faruqui15sparse} use sparse coding on conventional dense word embeddings in order to obtain sparse, higher dimensional and more interpretable vector spaces. Motivated by the success of neural architectures, deploying a sparse auto-encoder for pre-trained dense word embeddings is proposed in \cite{subramanian18spine} in order to improve interpretability. Instead of using sparse transformations as in the abovementioned studies, several other studies focused on learning orthogonal transformations that will preserve the internal semantic information and high performance of the original dense embedding space. In \cite{zobnin17rotations}, interpretability is taken as the tightness of clustering along individual embedding dimensions and orthogonal transformations are utilized to improve it. However, \cite{zobnin17rotations} has also shown that based on this definition for interpretability, total interpretability of an embedding is kept constant under any orthogonal transformation and it can only be redistributed across the dimensions. \cite{park17rotated} investigated rotation algorithms based on exploratory factor analysis (EFA) in order to improve interpretability while preserving the performance. \cite{dufter19ultradense} proposed a method to learn an orthogonal transformation matrix that will align a given linguistic signal in the form of a word group to an embedding dimension providing an interpretable subspace. In their work, they demonstrate their method for a one dimensional subspace. However, it is not clear how well the proposed method can generalize for a larger dimensional subspace (ideally the entire embedding space). In \cite{senel18semanticStructure} a transformation based on Bhattacharya distance and SEMCAT categories is proposed to obtain an interpretable embedding space. In that study, also an automated metric was proposed to quantitatively measure the degree of interpretability already present in the embedding vector spaces. Taking a different approach, \cite{herbelot15modelTheoretic} proposed a method to map dense word embeddings to a \textit{model-theoretic space} where the dimensions correspond to real world features elicited from human participants.

Following a separate line of work based on the research on topic modelling domain, \cite{panigrahi19word2sense} proposed an LDA (Latent Dirichlet Allocation) based generative model to extract different senses for words from a corpus. They also proposed a method to learn sparse interpretable word embedding, called Word2Sense, based on the obtained sense distributions. Several other studies also focused on associations between word embedding models and the topic modelling methods, \cite{liu15topical,moody16mixing,das15gaussian,shi17jointly}. They make use of LDA based models to obtain word topics to be integrated to word embeddings. It should be noted that this literature is also related in the sense that topic modelling may be used to improve the procedures for extracting the word-groups representing the concepts assigned to the embedding dimensions.

Most of the interpretability-related previous work mentioned above, except \cite{fyshe14interpretable}, \cite{senel18semanticStructure} and our proposed method, do not need external resources, utilization of which has both advantages and disadvantages. Methods that do not use external information require fewer resources but they also lack the aid of information extracted from these resources.

\section{Problem Description} \label{sec:problem_description}

For the task of unsupervised word embedding extraction, we operate on a discrete collection of lexical units (words) $u_i \in {{V}}$ that is part of an input corpus ${{C}} = \{u_i\}_{i\geq1} $, with number of tokens $\card{{C}}$, sourced from a vocabulary ${{V}} = \{w_1, \ldots, w_V \}$ of size $V$.\footnote{
We represent vectors (matrices) by bold lower (upper) case letters. For a vector $\vec{a}$ (a matrix $\matr{A}$), $\vec{a}^T$ ($\matr{A}^T$) is the transpose. $\norm{\vec{a}}$ stands for the Euclidean norm. For a set $S$, $\card{S}$ denotes the cardinality. $\indicator_{x \in S}$ is the indicator variable for the inclusion ${x \in S}$, evaluating to 1 if satisfied, 0 otherwise.
} In the setting of distributional semantics, the objective of a word embedding algorithm is to maximize some aggregate utility over the entire corpus so that some measure of ``closeness'' is maximized for pairs of vector representations $(\vec{w}_i, \vec{w}_j)$ for words which, on the average, appear in proximity to one another. In the GloVe algorithm \cite{pennington14glove}, which we base our proposed method upon, the following objective function is considered:
\begin{equation}
J = \sum_{i,j=1}^{V} f(X_{ij}) \left(\vec{w}_i^T\vec{\tilde{w}}_j + b_i + \tilde{b}_j -\log{}X_{ij}\right)^2.
\label{eq:glove}
\end{equation}

In \eqref{eq:glove}, $\vec{w}_i \in \Real^D$ and $\vec{\tilde{w}}_j \in \Real^D$ stand for word and context vector representations, respectively, for words $w_i$ and $w_j$, while $X_{ij}$ represents the (possibly weighted) cooccurrence count for the word pair $(w_i, w_j)$. Intuitively, \eqref{eq:glove} represents the requirement that if some word $w_i$ occurs often enough in the context (or vicinity) of another word $w_j$, then the corresponding word representations should have a large enough inner product in keeping with their large $X_{ij}$ value, up to some bias terms $b_i, \tilde{b}_j$; and vice versa.
$f(\cdot)$ in \eqref{eq:glove} is used as a discounting factor that prohibits rare cooccurrences from disproportionately influencing the resulting embeddings.

The objective \eqref{eq:glove} is minimized using stochastic gradient descent by iterating over the matrix of cooccurrence records $[X_{ij}]$. In the GloVe algorithm, for a given word $w_i$, the final word representation is taken to be the average of the two intermediate vector representations obtained from \eqref{eq:glove}; i.e, $(\vec{w}_i + \vec{\tilde{w}}_j)/2$. In the next section, we detail the enhancements made to \eqref{eq:glove} for the purposes of enhanced interpretability, using the aforementioned framework as our basis. 

\section{Imparting Interpretability} \label{sec:infl}

\begin{figure}[!t]
   \centering
	\includegraphics[width=10cm]{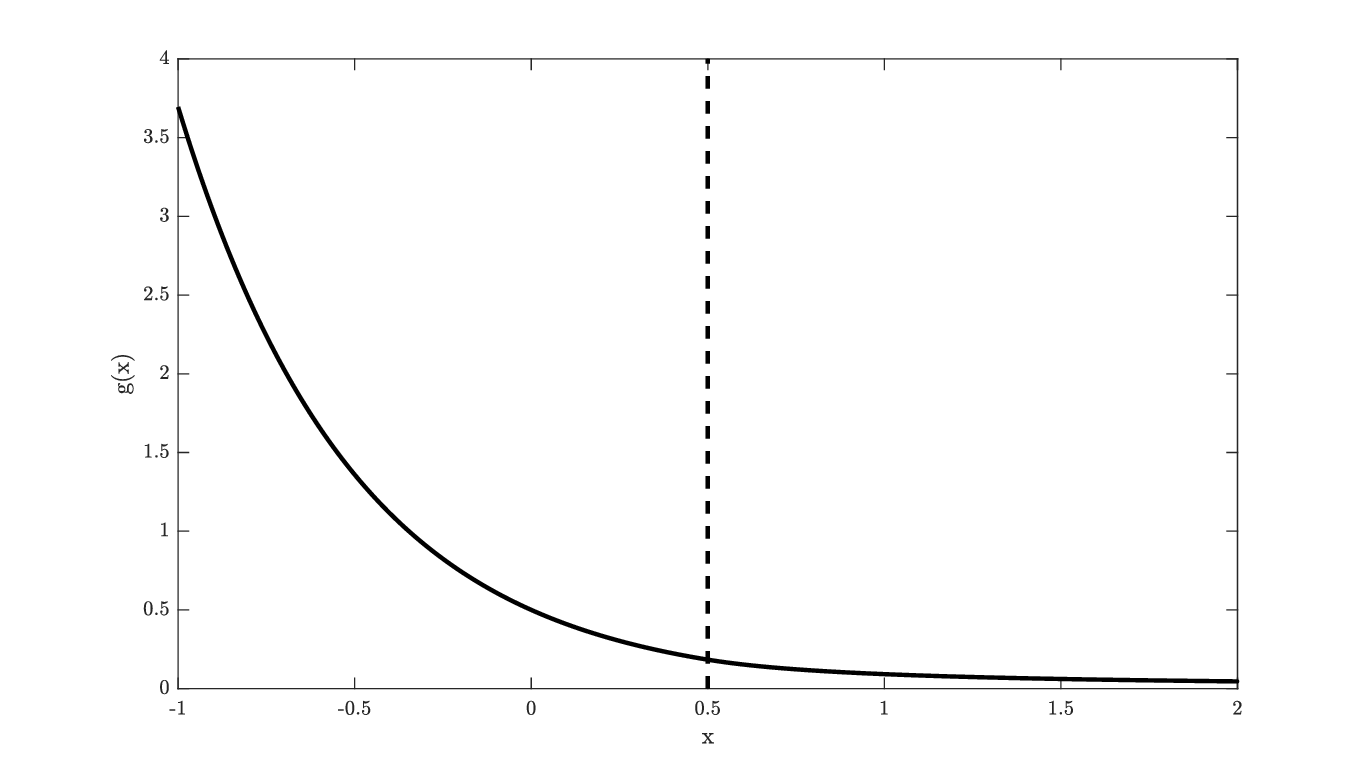}
	\caption{Function $g$ in the additional cost term.}
	\label{fig:g_of_x}
\end{figure}

Our approach falls into a joint-learning framework where the distributional information extracted from the corpus is allowed to fuse with the external lexicon-based information. An external resource in which words are primarily grouped together based on human judgments and in which the entire semantic space is represented as much as possible is needed. We have chosen to use Roget’s Thesaurus both due to its being one of the earliest examples of its kind, but also its status of being continuously updated for modern words. \emph{Word-groups} extracted from Roget's Thesaurus are directly mapped to individual dimensions of word embeddings. Specifically, the vector representations of words that belong to a particular group are encouraged to have deliberately increased values in a particular dimension that corresponds to the word-group under consideration. This can be achieved by modifying the objective function of the embedding algorithm to partially influence vector representation distributions across their dimensions over an input vocabulary. To do this, we propose the following modification to the GloVe objective given in Eq.~\eqref{eq:glove}:
\begin{align}
\begin{split}
	J  =  & \sum_{i,j=1}^{\card{{V}}} f(X_{ij})\Bigg[ \left(\vec{w}_i^T\vec{\tilde{w}}_j + b_i + \tilde{b}_j -\log{}X_{ij}\right)^2 \\
	      & +\: k\left(\sum_{l=1}^{D} \indicator_{i\in{}F_l} \: g(\vec{w}_{i,l}) + \sum_{l=1}^{D} \indicator_{j \in F_l} \: g(\vec{\tilde{w}}_{j,l})  \right) \Bigg] \label{eq:proposed}
\end{split}
\end{align}
In Eq.~\eqref{eq:proposed}, $F_l$ denotes the indices for the elements of the $l$th concept word-group which we wish to assign in the vector dimension~$l = 1, \ldots, D$. The objective in Eq.~\eqref{eq:proposed} is designed as a mixture of two individual cost terms: the original GloVe cost term along with a second term that encourages embedding vectors of a given concept word-group to achieve deliberately increased values along an associated dimension $l$. The relative weight of the second term is controlled by the parameter $k$. The simultaneous minimization of both objectives ensures that words that are similar to, but not included in, one of these concept word-groups are also ``nudged'' towards the associated dimension $l$. The trained word vectors are thus encouraged to form a distribution where the individual vector dimensions align with certain semantic concepts represented by a collection of concept word-groups, one assigned to each vector dimension. To facilitate this behavior, Eq.~\eqref{eq:proposed} introduces a monotone decreasing function $g(\cdot)$ defined as
\begin{equation}
\label{eq:g}
g(x) =
\begin{cases}
\cfrac{1}{2}\: \exp\left(-{2x}\right) & \text{for } x<0.5, \\[10pt]
\cfrac{1}{(4e)x} & \text{otherwise},
\end{cases}
\end{equation}
which serves to increase the total cost incurred if the value of the $l$th dimension for the two vector representations $\vec{w}_{i,l}$ and $\vec{\tilde{w}}_{j,l}$ for a concept word $\vec{w}_i$ with $i \in F_l$ fails to be large enough. Although different definitions for $g(x)$ are possible, we observed after several experiments that this piece-wise definition provides decent push for the words in the word-groups. $g(x)$ is graphically shown in Fig.~\ref{fig:g_of_x} and we will also present further analysis and experiments regarding the effects of different forms of $g(x)$ in Section~\ref{g_selection}.

The objective Eq.~\eqref{eq:proposed} is minimized using stochastic gradient descent over the cooccurrence records $\{X_{ij}\}_{i,j = 1}^{\card{{V}}}$. Intuitively, the terms added to Eq.~\eqref{eq:proposed} in comparison with Eq.~\eqref{eq:glove} introduce the effect of selectively applying a positive step-type input to the original descent updates of Eq.~\eqref{eq:glove} for concept words along their respective vector dimensions, which influences the dimension value in the positive direction. The parameter $k$ in Eq.~\eqref{eq:proposed} allows for the adjustment of the magnitude of this influence as needed.

In the next section, we demonstrate the feasibility of this approach by experiments with an example collection of concept word-groups extracted from Roget's Thesaurus. Before moving on, we would like to first comment on the discovery of the “ultimate” categories. Questions like “What are the intrinsic and fundamental building blocks of the entire semantic space?” and “What should be the corresponding categories?” are important questions in linguistics and natural language processing, and above all, in philosophy. Determining them without human intervention and with unsupervised means is an open problem. Roget’s Thesaurus can be seen as a manual attempt for answering this question. The methodology behind it is exhaustive. Based on the premise that there should be some word in a language for any material or immaterial thing known to humans, Roget’s Thesaurus conceptualizes all the words within a tree structure with hierarchical categories. So, taking it as starting point to construct the categories and concept word-groups is a logical option. One can readily use any other external lexical resource as long as the groupings of words are not arbitrary (which is the case in all thesauruses) or leverage topic modelling methods to form the categories. It is also obvious that some sort of supervision is better in terms of reaching the “ultimate” categories than unsupervised approaches. On the other side, however, unsupervised methods have the advantage on not depending on external resources. This leads to the question of how supervised and unsupervised approaches
compare. For that reason, in the next section, we also quantitatively compare our method against supervised and unsupervised projection based simple baselines for concept word-groups extraction.

\section[Experiments and Results]{Experiments and Results\footnote{All necessary source codes to reproduce the experiments in this section are available at https://github.com/koc-lab/imparting-interpretability}}

\begin{table}[!b]
	\caption{\protect\centering Sample concepts and their associated word-groups from Roget's Thesaurus}
	\centering
	\label{table:roget_sample_words}
	\begin{tabular}{ccccc}
		\hline 
		\multirow{2}{*}{\tt MANKIND } & \multirow{2}{*}{ \tt BUSINESS } & {\tt SIMPLE} & \multirow{2}{*}{\tt CONDUCT } & \multirow{2}{*}{ \tt ARRIVAL} \\
		& & {\tt QUANTITY} & & \\
		\hline
		one & living & size & government & land \\
		population & work & way & life & home \\\
		people & line & point & game & light \\\
		world & place & force & role & airport \\\
		state & service & station & race & return \\\
		family & role & range & record & come \\\
		national & race & standard & process & complete \\\
		public & office & rate & business & port \\\
		party & act & stage & career & hit \\\
		million & case & mass & campaign & meeting \\\
		\vdots & \vdots & \vdots & \vdots & \vdots \\
		\hline 
	\end{tabular}
\end{table}

\begin{table}[!b]
	\caption{\protect\centering GloVe Parameters}
	\label{table:glove_params}
	\centering
	\begin{tabular}{lr}
		\hline 
		\verb|VOCAB_MIN_COUNT| & 65\\\
		\verb|ALPHA| & 0.75 \\\
		\verb|WINDOW_SIZE| & 15 \\\
		\verb|VECTOR_SIZE| & 300 \\\
		\verb|X_MAX| & 75 \\
		\hline 
	\end{tabular}
\end{table}

\label{sec:exp}
We first identified 300 concepts, one for each dimension of the 300-dimensional vector representation, by employing Roget's Thesaurus. This thesaurus follows a tree structure which starts with a \emph{Root} node that contains all the words and phrases in the thesaurus. The root node is successively split into \emph{Classes} and \emph{Sections}, which are then (optionally) split into \emph{Subsections} of various depths, finally ending in \emph{Categories}, which constitute the smallest unit of word/phrase collections in the structure. The actual words and phrases descend from these \textit{Categories}, and make up the leaves of the tree structure. We note that a given word typically appears in multiple categories corresponding to the different senses of the word. We constructed concept word-groups from Roget's Thesaurus as follows: We first filtered out the multi-word phrases and the relatively obscure terms from the thesaurus. The obscure terms were identified by checking them against a vocabulary extracted from Wikipedia. We then obtained 300 word-groups as the result of a \emph{partitioning} operation applied to the subtree that ends with \emph{categories} as its leaves. The partition boundaries, hence the resulting word-groups, can be chosen in many different ways. In our proposed approach, we have chosen to determine this partitioning by traversing this tree structure from the root node in breadth-first order, and by employing a parameter $\lambda$ for the maximum \emph{size} of a node. Here, the size of a node is defined as the number of unique words that ever-descend from that node. During the traversal, if the size of a given node is less than this threshold, we designate the words that ultimately descend from that node as a concept word-group. Otherwise, if the node has children, we discard the node, and queue up all its children for further consideration. If this node does not have any children, on the other hand, the node is truncated to $\lambda$ elements with the highest frequency-ranks, and the resulting words are designated as a concept word-group. The algorithm for extracting concept word-groups from Roget's Thesaurus is also given in pseudo code form as Algorithm~\ref{algorithmRoget}. We note that the choice of $\lambda$ greatly affects the resulting collection of word-groups: Excessively large values result in few word-groups that greatly overlap with one another, while overly small values result in numerous tiny word-groups that fail to adequately represent a concept. We experimentally determined that a $\lambda$ value of 452 results in the most healthy number of relatively large word-groups (113 groups with size $\geq$ 100), while yielding a preferably small overlap amongst the resulting word-groups (with average overlap size not exceeding 3 words). A total of 566 word-groups were thus obtained. 259 smallest word-groups (with size $<$ 38) were discarded to bring down the number of word-groups to 307. Out of these, 7 groups with the lowest median frequency-rank were further discarded, which yields the final 300 concept word-groups used in the experiments. We present some of the resulting word-groups in Table~\ref{table:roget_sample_words}.\footnote{All the vocabulary lists, concept word-groups and other material necessary to reproduce this procedure is available at https://github.com/koc-lab/imparting-interpretability}

\vspace{.5cm}
\begin{algorithm}[h]
     word\_groups = [ ]\\
     nodes = [ ]\\
     nodes.add(root\_node)\\
     \While{length of nodes $>$ 0}{
        large\_nodes = nodes\\
        \For{node in large\_nodes}{
            \eIf{node.size() $>$ $\lambda$}{
                children\_nodes = node.get\_children()\\
                \eIf{length of children\_nodes $>$ 0}{
                    nodes.add(children\_nodes)
                }{
                    word\_groups.add(node.truncate\_to\_lambda())
                }
            }{
                word\_groups.add(node)
            }
            nodes.remove(node)
        }
    }
    \caption{Algorithm for extracting concept word-groups} 
    \label{algorithmRoget}
\end{algorithm} 
\vspace{.5cm}

By using the concept word-groups, we have trained the GloVe algorithm with the proposed modification given in Section~\ref{sec:infl} on a snapshot of English Wikipedia consisting of around 1.1B tokens, with the stop-words filtered out. Using the parameters given in Table~\ref{table:glove_params}, this resulted in a vocabulary size of 287,847. For the weighting parameter in Eq.~\eqref{eq:proposed}, we used a value of $k = 0.1$ whose effect is analysed in detail in Section~\ref{k_selection}. The algorithm was trained over 20 iterations. The GloVe algorithm without any modifications was also trained with the same parameters. In addition to the original GloVe algorithm, we compare our proposed method with previous studies that aim to obtain interpretable word vectors. We train the \textit{improved projected gradient} model proposed in \cite{luo15online} to obtain word vectors (called OIWE-IPG) using the same corpus we use to train GloVe and our proposed method. For the Word2Sense method \cite{panigrahi19word2sense}, we use 2250 dimensional pretrained embeddings for comparisons instead of training the algorithm on the same corpus used for the other methods due to very slow training of the model on our hardware\footnote{Pretrained Word2Sense model has advantage over our proposed method and the other alternatives due to being trained on a nearly 3 times larger corpus (around 3B tokens).}. Using the methods proposed in \cite{faruqui15sparse,park17rotated,subramanian18spine} on our baseline GloVe embeddings, we obtain SOV, SPINE and Parsimax (orthogonal) word representations, respectively. We train all the models with the proposed parameters. However, in \cite{subramanian18spine}, the authors show results for a relatively small vocabulary of 15,000 words. When we trained their model on our baseline GloVe embeddings with a large vocabulary of size 287,847, the resulting vectors performed significantly poor on word similarity tasks compared to the results presented in their paper. In addition to these alternatives, we also compare our method against two simple projection based baselines. Specifically, we construct two new embedding spaces by projecting the original GloVe embeddings onto i) randomly sampled 300 different tokens ii) average vectors of the words for the 300 word-groups extracted from Roget's Thesaurus. We evaluate the interpretability of the resulting embeddings qualitatively and quantitatively. We also test the performance of the embeddings on word similarity and word analogy tests as well as on a downstream classification task. 

In our experiments, vocabulary size is close to 300,000 while only 16,242 unique words of the vocabulary are present in the concept groups. Furthermore, only dimensions that correspond to the concept group of the word will be updated due to the additional cost term. Given that these concept words can belong to multiple concept groups (2 on average), only 33,319 parameters are updated. There are 90 million individual parameters present for the 300,000 word vectors of size 300. Of these parameters, only approximately 33,000 are updated by the additional cost term. For the interpretability evaluations, we restrict the vocabulary to the most frequent 50,000 words\footnote{Since Word2Sense embeddings has a different vocabulary, we first filter and sort it based on our vocabulary.} except Fig.~\ref{fig:sim} where we only use most frequent 1000 words for clarity of the plot.

\begin{table}
	\renewcommand{\arraystretch}{1.3}
	\caption{\protect\centering Words with largest dimension values for the proposed algorithm}
	\label{table:effect_on_top_words}
	\centering
	\resizebox{\columnwidth}{!}{%
	\begin{tabular}{ccccccccc}
		\hline 
		\multirow{2}{*}{\tt GOVERNMENT}  &  \multirow{2}{*}{\tt CHOICE}  &  \multirow{2}{*}{\tt BOOK}  &  \multirow{2}{*}{\tt NEWS}  &  {\tt PROPERTY}   \\
		 & & & & {\tt IN GENERAL}\\ \hline
		\forced republic  &  \forced poll  &  \forced editor  &  \forced radio  &  \forced lands  \\
		\forced province  &  \notforced shortlist  &  \forced publisher  &  \forced news  &  \forced land  \\
		\notforced provinces  &  \forced vote  &  \forced magazine  &  \notforced tv  &  \forced ownership  \\
		\forced government  &  \forced selection  &  \forced writer  &  \notforced broadcasting  &  \forced possession  \\
		\forced administration  &  \notforced televoting  &  \forced author  &  \notforced broadcast  &  \forced assets  \\
		\forced prefecture  &  \forced preference  &  \notforced hardcover  &  \notforced broadcasts  &  \forced acquired  \\
		\forced governor  &  \forced choosing  &  \notforced paperback  &  \notforced simulcast  &  \forced property  \\
		\forced county  &  \forced choose  &  \notforced books  &  \notforced channel  &  \forced acres  \\
		\forced monarchy  &  \forced choice  &  \forced page  &  \forced television  &  \forced estate  \\
		\forced region  &  \forced chosen  &  \forced press  &  \notforced cnn  &  \forced lease  \\
		\forced territory  &  \forced elect  &  \notforced publishing  &  \notforced jazeera  &  \forced inheritance  \\
		\notforced autonomous  &  \forced list  &  \notforced edited  &  \notforced fm  &  \forced manor  \\
		\forced administrative  &  \forced election  &  \forced volume  &  \notforced programming  &  \notforced holdings  \\
		\forced minister  &  \forced select  &  \forced encyclopedia  &  \notforced bbc  &  \notforced ploughs  \\
		\forced senate  &  \forced preferential  &  \notforced published  &  \notforced newscast  &  \notforced estates  \\
		\notforced districts  &  \forced option  &  \notforced publications  &  \notforced simulcasts  &  \forced owner  \\
		\notforced democratic  &  \notforced voters  &  \forced bibliography  &  \notforced syndicated  &  \forced feudal  \\
		\forced legislature  &  \notforced ballots  &  \forced periodical  &  \forced media  &  \notforced heirs  \\
		\notforced abolished  &  \notforced votes  &  \forced publication  &  \forced reporter  &  \forced freehold  \\
		\forced presidency  &  \notforced sssis  &  \forced essayist  &  \notforced cbs  &  \forced holding  \\
		\vdots  &  \vdots  &  \vdots  &  \vdots  &  \vdots  \\ \hline 

		\multirow{2}{*}{\tt TEACHING}  &  {\tt NUMBERS IN}   &  \multirow{2}{*}{\tt PATERNITY}  &  \multirow{2}{*}{\tt WARFARE}  &  \multirow{2}{*}{\tt FOOD}  \\ 
		 & {\tt THE ABSTRACT} & & & \\ \hline
		\forced curriculum  &  \notforced integers  &  \forced family  &  \forced battle  & \forced meal \\
		\forced exam  &  \notforced polynomial  &  \forced paternal  &  \forced war  & \notforced dishes  \\
		\forced training  &  \forced integer  &  \forced maternal  &  \notforced battles  & \forced bread \\
		\forced school  &  \notforced polynomials  &  \forced father  &  \forced combat  & \notforced eaten \\
		\notforced students  &  \forced logarithm  &  \forced grandfather  &  \forced military  & \forced dessert \\
		\notforced toefl  &  \forced modulo  &  \forced grandmother  &  \forced warfare  & \notforced cooked \\
		\notforced exams  &  \forced formula  &  \forced mother  &  \forced fighting  & \notforced  foods \\
		\forced teaching  &  \notforced coefficients  &  \forced ancestry  &  \forced battlefield  & \forced dish \\
		\notforced schools  &  \forced multiplication  &  \notforced son  &  \notforced guerrilla  & \forced food \\
		\forced education  &  \forced finite  &  \notforced hemings  &  \notforced fought  & \forced meat \\
		\forced teach  &  \notforced logarithms  &  \forced ancestor  &  \forced campaign  & \forced eating \\
		\notforced karate  &  \forced algebra  &  \notforced patrilineal  &  \forced fight  & \forced cuisine \\
		\forced taught  &  \notforced integrals  &  \notforced daughter  &  \notforced insurgency  & \forced beverage \\
		\notforced courses  &  \notforced primes  &  \notforced grandson  &  \forced armed  & \forced soup \\
		\notforced civics  &  \forced divisor  &  \forced descent  &  \forced tactics  & \forced snack \\
		\forced instruction  &  \forced compute  &  \forced house  &  \forced operations  & \forced pork \\
		\notforced syllabus  &  \forced arithmetic  &  \notforced parents  &  \notforced army  & \forced eat \\
		\forced test  &  \forced algorithm  &  \notforced descendant  &  \notforced mujahideen  & \forced wine \\
		\notforced examinations  &  \notforced theorem  &  \notforced grandparents  &  \notforced armies  & \forced beef \\
		\notforced instructor  &  \notforced quadratic  &  \forced line  &  \notforced soldiers  & \notforced fried \\
		\vdots  &  \vdots  &  \vdots  &  \vdots   &  \vdots   \\ \hline
			
	\end{tabular}
	}
\end{table}

\subsection{Qualitative Evaluation for Interpretability}

\begin{sidewaysfigure*}
  \includegraphics[width=1\textwidth]{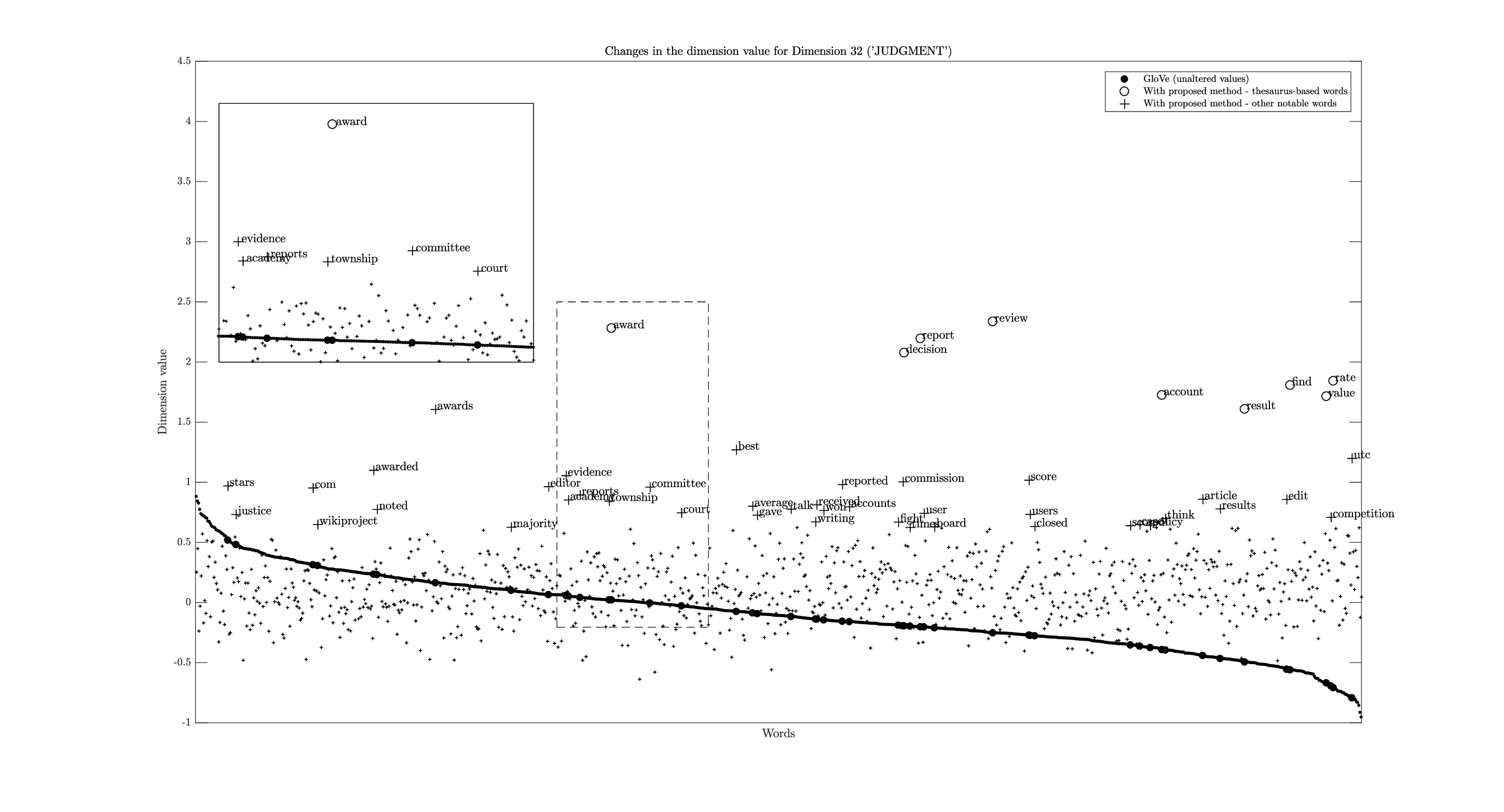}
  \caption{Most frequent 1000 words sorted according to their values in the $32^{nd}$ dimension of the original GloVe embedding are shown with ``$\bullet$" markers. ``$\circ$" and ``$+$" markers show the values of the same words for the $32^{nd}$ dimension of the embedding obtained with the proposed method where the dimension is \textit{aligned\/} with the concept {\tt JUDGMENT}. Words with ``$\circ$" markers are contained in the concept {\tt JUDGMENT} while words with ``$+$" markers are not contained.}
  \label{fig:sim} 
\end{sidewaysfigure*}

In Fig.~\ref{fig:sim}, we demonstrate the particular way in which the proposed algorithm Eq.~\eqref{eq:proposed} influences the vector representation distributions. Specifically, we consider, for illustration, the $32^{nd}$ dimension values for the original GloVe algorithm and our modified version, restricting the plots to the top-1000 words with respect to their frequency ranks for clarity of presentation. In Fig.~\ref{fig:sim}, the words in the horizontal axis are sorted in descending order with respect to the values at the $32^{nd}$ dimension of their word embedding vectors coming from the original GloVe algorithm. The dimension values are denoted with ``$\bullet$" and ``$\circ$"/``$+$" markers for the original and the proposed algorithms, respectively. Additionally, the top-50 words that achieve the greatest $32^{nd}$ dimension values among the considered 1000 words are emphasized with enlarged markers, along with text annotations. In the presented simulation of the proposed algorithm, the $32^{nd}$ dimension values are encoded with the concept \verb|JUDGMENT|, which is reflected as an increase in the dimension values for words such as \verb|committee|, \verb|academy|, and \verb|article|. We note that these words (denoted by $+$) are \emph{not\/} part of the pre-determined word-group for the concept \verb|JUDGMENT|, in contrast to words such as \verb|award|, \verb|review| and \verb|account| (denoted by $\circ$) which are. This implies that the increase in the corresponding dimension values seen for these words is attributable to the joint effect of the first term in Eq.~\eqref{eq:proposed} which is inherited from the original GloVe algorithm, in conjunction with the remaining terms in the proposed objective expression Eq.~\eqref{eq:proposed}. This experiment illustrates that the proposed algorithm is able to impart the concept of \verb|JUDGMENT| on its designated vector dimension above and beyond the supplied list of words belonging to the concept word-group for that dimension. It should also be noted that the majority of the words in Fig.~\ref{fig:sim} are denoted by ``$+$", which means that they are \emph{not\/} part of the pre-determined word-groups and are semi-supervisedly imparted.


We also present the list of words with the greatest dimension value for the dimensions 11, 13, 16, 31, 36, 39, 41, 43 and 79 in Table~\ref{table:effect_on_top_words}. These dimensions are aligned/imparted with the concepts that are given in the column headers. In Table~\ref{table:effect_on_top_words}, the words that are given with regular font denote the words that exist in the corresponding word-group obtained from Roget's Thesaurus (and are thus explicitly forced to achieve increased dimension values), while emboldened words denote the words that achieve increased dimension values by virtue of their co-occurrence statistics with the thesaurus-based words (indirectly, without being explicitly forced). This again illustrates that a semantic concept can indeed be coded to a vector dimension provided that a sensible lexical resource is used to guide semantically related words to the desired vector dimension via the proposed objective function in Eq.~\eqref{eq:proposed}. Even the words that do not appear in, but are semantically related to, the word-groups that we formed using Roget's Thesaurus, are indirectly affected by the proposed algorithm. They also reflect the associated concepts at their respective dimensions even though the objective functions for their particular vectors are not modified. This point cannot be overemphasized. Although the word-groups extracted from Roget's Thesaurus impose a degree of supervision to the process, the fact that the remaining words in the entire vocabulary are also indirectly affected makes the proposed method a semi-supervised approach that can handle words that are not in these chosen word-groups. A qualitative example of this result can be seen in the last column of Table~\ref{table:effect_on_top_words}. It is interesting to note the appearance of words such as \verb|guerilla|, \verb|insurgency|, \verb|mujahideen|, \verb|Wehrmacht| and \verb|Luftwaffe| in addition to the more obvious and straightforward \verb|army|, \verb|soldiers| and \verb|troops|, all of which are not present in the associated word-group \verb|WARFARE|.

Most of the dimensions we investigated exhibit similar behavior to the ones presented in Table~\ref{table:effect_on_top_words}. Thus generally speaking, we can say that the entries in Table~\ref{table:effect_on_top_words} are representative of the great majority. However, we have also specifically looked for dimensions that make less sense and determined a few such dimensions which are relatively less satisfactory. These less satisfactory examples are given in Table~\ref{table:effect_on_top_words_bad}. These examples are also interesting in that they shed insight into the limitations posed by polysemy and existence of very rare outlier words. 

\begin{table}
	\renewcommand{\arraystretch}{1.3}
	\caption{\protect\centering Words with largest dimension values for the proposed algorithm - Less Satisfactory Examples}
	\label{table:effect_on_top_words_bad}
	\centering
	\begin{tabular}{cccc}
		\hline
		\verb|MOTION| & \verb|TASTE| & \verb|REDUNDANCY| & \verb|FEAR| \\ \hline
		\notforced nektonic & \forced polish & \notforced eusebian & \forced horror \\ 
		\forced rate & \forced classical & \forced margin & \forced fear \\
		\forced mobile & \forced taste & \forced drug & \forced dread \\ 
		\forced movement & \forced culture & \notforced arra & \forced trembling \\ 
		\forced motion & \forced corinthian & \forced overflow & \forced scare \\ 
		\forced evolution & \notforced przeworsk & \forced overdose & \forced terror \\ 
		\forced gait & \forced artistic & \forced extra & \forced panic \\ 
		\forced velocity & \forced judge & \forced excess & \forced anxiety \\ 
		\notforced novokubansk & \forced aesthetic & \forced bonus & \notforced $\phi$\textit{ò}$\beta$\textit{o}$\varsigma$ \\ 
		\notforced brownian & \forced amateur & \notforced synaxarion & \notforced phobia \\ 
		\forced port & \forced critic & \forced load & \forced fright \\ 
		\forced flow & \notforced kraków & \notforced padding & \forced terrible \\ 
		\forced gang & \forced elegance & \forced crowd & \forced frighten \\ 
		\forced roll & \forced aesthetics & \forced redundancy & \forced pale \\ 
		\forced stride & \notforced plaquemine & \forced overrun & \notforced vacui \\ 
		\forced run & \forced judgment & \notforced boilerplate & \forced haunt \\ 
		\forced kinematics & \forced connoisseur & \forced excessive & \forced afraid \\ 
		\forced stream & \notforced katarzyna & \notforced $\tau\iota\tau\lambda$\textit{o}$\iota$ & \forced fearful \\ 
		\forced walk & \notforced cucuteni & \forced lavish & \forced frightened \\ 
		\forced drift & \notforced warsaw & \forced gorge & \forced shaky \\ 
		\vdots & \vdots & \vdots & \vdots \\ \hline
	\end{tabular}
\end{table}

\subsection{Quantitative Evaluation for Interpretability}

One of the main goals of this study is to improve the interpretability of dense word embeddings by aligning the dimensions with predefined concepts from a suitable lexicon. A quantitative measure is required to reliably evaluate the achieved improvement. One of the methods to measure the interpretability is the word intrusion test \cite{chang09wordintrusion} where manual evaluations from multiple human evaluators for each embedding dimension are used. Deeming this manual method expensive to apply, \cite{senel18semanticStructure} introduced a semantic category-based approach and category dataset (SEMCAT) to automatically quantify interpretability. We use both of these approaches to quantitatively verify our proposed method in the following two subsections.

\subsubsection{Automated Evaluation for Interpretability} Specifically, we apply a modified version of the approach presented in \cite{senel18trInterpret} in order to consider possible sub-groupings within the categories in SEMCAT\footnote{Please note that the usage of ``category" here in the setting of SEMCAT should not be confused with the ``categories" of Roget's Thesaurus.}. Interpretability scores are calculated using Interpretability Score (IS) as given below:

\vspace*{-0.3cm}
\begin{equation} \label{eq:interpretability}
\begin{split}
&IS^+_{i,j} = \max_{n_{min} \leq n \leq n_j } \frac{|S_j \cap V^+_i(\lambda \times n)|}{n} \times 100  \\
&IS^-_{i,j} = \max_{n_{min} \leq n \leq n_j } \frac{|S_j \cap V^-_i(\lambda \times n)|}{n} \times 100  \\
&IS_{i,j} = \max(IS^+_{i,j}, IS^-_{i,j}) \\
&IS_{i} = \max_{j} IS_{i,j}, ~~ IS = \frac{1}{D}\sum\limits_{i=1}^D IS_{i}
\end{split}
\end{equation} 

In Eq.~\eqref{eq:interpretability}, $IS^+_{i,j}$ and $IS^-_{i,j}$ represents the interpretability scores in the positive and negative directions of the $i^{th}$ dimension ($i \in \{1,2,...,D\}$, $D$ number of dimensions in the embedding space) of word embedding space for the $j^{th}$ category ($j \in \{1,2,...,K\}$, $K$ is number of categories in SEMCAT, $K=110$) in SEMCAT respectively. $S_j$ is the set of words in the $j^{th}$ category in SEMCAT and $n_j$ is the number of words in $S_j$.  $n_{min}$ corresponds to the minimum number of words required to construct a semantic category (i.e.\ represent a concept). $V_i(\lambda \times n_j)$ represents the set of $\lambda \times n_j$ words that have the highest ($V_i^+$) and lowest ($V_i^-$) values in $i^{th}$ dimension of the embedding space. $\cap$ is the intersection operator and $|.|$ is the cardinality operator (number of elements) for the intersecting set. In Eq.~\eqref{eq:interpretability}, $IS_{i}$ gives the interpretability score for the $i^{th}$ dimension and $IS$ gives the average interpretability score of the embedding space. 

\begin{figure}[!t]
    \centering
     \includegraphics[width=9.5cm]{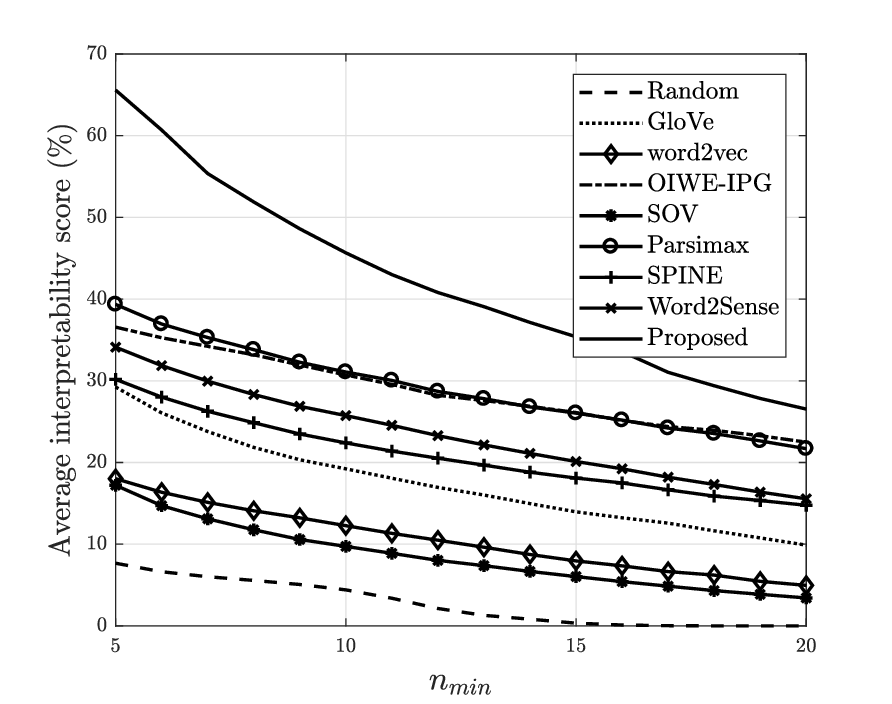}  
    \newline
    \newline
    \caption{Interpretability scores averaged over 300 dimensions for the original GloVe method, the proposed method, and five alternative methods along with a randomly generated baseline embedding for $\lambda = 5$.}
    \label{fig:interpretability}
\end{figure}

Fig.~\ref{fig:interpretability} presents the measured average interpretability scores across dimensions for original GloVe embeddings, for the proposed method and for the other five methods we compare, along with a randomly generated embedding. Results are calculated for the parameters $\lambda = 5$ and $n_{min} \in \{5,6,\dots,20\}$. Our proposed method significantly improves the interpretability for all $n_{min}$ compared to the original GloVe approach and it outperforms all the alternative approaches by a large margin especially for lower $n_{min}$.

The proposed method and interpretability measurements are both based on utilizing concepts represented by word-groups. Therefore it is expected that there will be higher interpretability scores for some of the dimensions for which the imparted concepts are also contained in SEMCAT. However, by design, word groups that they use are formed by using different sources and are independent. Interpretability measurements use SEMCAT while our proposed method utilizes Roget's Thesaurus.

\subsubsection{Human Evaluation for Interpretability: Word Intrusion Test}\label{sec:intrusion}
Although measuring interpretability of imparted word embeddings with SEMCAT gives successful results, making another test involving human judgment surely enhances reliability of imparted word embeddings. One of the tests that includes human judgment for interpretability is the word intrusion test, \cite{chang09wordintrusion}. Word intrusion test is a multiple choice test where each choice is a separate word. Four of these words are chosen among the words whose vector values at a specific dimension are high and one is chosen from the words whose vector values are low at that specific dimension. This word is called an intruder word. If the participant can distinguish the intruder word from others, it can be said that this dimension is interpretable. If the underlying word embeddings are interpretable across the dimension, the intruder words can be easily found by human evaluators.

In order to increase the reliability of the test, we both used imparted and original GloVe embeddings for comparison. For each dimension of both embeddings, we prepare a question. We shuffled the questions in random order so that participant cannot know which question comes from which embedding. In total, there are 600 questions (300 GloVe + 300 imparted GloVe) with five choices for each.\footnote{Questions in this word intrusion test are available at https://github.com/koc-lab/imparting-interpretability} We apply the test on five participants. Results tabulated at Table~\ref{table:intrusiontest} show that our proposed method significantly improves the interpretability by increasing the average correct answer percentage from $\sim 28 \%$ for baseline to $\sim 71 \%$ for our method.

\begin{table}
\centering
\caption{\protect\centering Word Intrusion Test Results: Correct Answers out of 300 Questions}
\begin{tabular}{lcc}
\hline
& GloVe & Imparted \\  \hline
Participants 1 to 5 & 80/88/82/78/97    & 212/170/207/229/242  \\
Mean/Std & 85/6.9 & 212/24.4 \\ \hline
\end{tabular}
\label{table:intrusiontest}
\end{table}

\subsection{Performance Evaluation of the Embeddings}

It is necessary to show that the semantic structure of the original embedding has not been damaged or distorted as a result of aligning the dimensions with given concepts, and that there is no substantial sacrifice involved from the performance that can be obtained with the original GloVe. To check this, we evaluate performances of the proposed embeddings on word similarity \cite{faruqui14communityEval} and word analogy \cite{mikolov13word2vec_a} tests. We also measure the performance on a downstream sentiment classification task. We compare the results with the original embeddings and the four alternatives excluding Parsimax \cite{park17rotated} since orthogonal transformations will not affect the performance of the original embeddings on these tests. 

Word similarity test measures the correlation between word similarity scores obtained from human evaluation (i.e. true similarities) and from word embeddings (usually using cosine similarity). In other words, this test quantifies how well the embedding space reflects human judgements in terms of similarities between different words. The correlation scores for 13 different similarity test sets and their averages are reported in Table~\ref{table:similarity_scores}. We observe that, let alone a reduction in performance, the obtained scores indicate an almost uniform improvement in the correlation values for the proposed algorithm, outperforming all the alternatives except Word2Vec baseline on average. Although Word2Sense performed slightly better on some of the test sets, it should be noted that it is trained on a significantly larger corpus. Categories from Roget's thesaurus are groupings of words that are similar in some sense which the original embedding algorithm may fail to capture. These test results signify that the semantic information injected into the algorithm by the additional cost term is significant enough to result in a measurable improvement. It should also be noted that scores obtained by SPINE is unacceptably low on almost all tests indicating that it has achieved its interpretability performance at the cost of losing its semantic functions.

\begin{table}
	\caption{\protect\centering Correlations for Word Similarity Tests}
	\label{table:similarity_scores}
	\centering
	\hspace{-0.35cm}
	\addtolength{\tabcolsep}{-2pt}
    \resizebox{\columnwidth}{!}{%
	\begin{tabular}{lccccccc}
		\hline 
		Dataset (EN-) & GloVe & Word2Vec & OIWE-IPG & SOV & SPINE & Word2Sense & Proposed \\
		\hline
		WS-353-ALL & $0.612$ & $0.7156$ &$0.634$ & $0.622$ & $0.173$ & $0.690$ & $0.657$\\
		SIMLEX-999 & $0.359$ & $0.3939$ &$0.295$ & $0.355$ & $0.090$ & $0.380$ & $0.381$ \\
		VERB-143   & $0.326$ & $0.4430$ &$0.255$ & $0.271$ & $0.293$ & $0.271$ & $0.348$ \\
		SimVerb-3500 & $0.193$ & $0.2856$ & $0.184$ & $0.197$ & $0.035$ & $0.234$ & $0.245$\\
		WS-353-REL & $0.578$ & $0.6457$ &$0.595$ & $0.578$ & $0.134$ & $0.695$ & $0.619$ \\
		RW-STANF.  & $0.378$ & $0.4858$ &$0.316$ & $0.373$ & $0.122$ & $0.390$ & $0.382$ \\
		YP-130     & $0.524$ & $0.5211$ &$0.353$ & $0.482$ & $0.169$ & $0.420$ & $0.589$ \\
		MEN-TR-3k  & $0.710$ & $0.7528$ &$0.684$ & $0.696$ & $0.298$ & $0.769$ & $0.725$ \\
		RG-65      & $0.768$ & $0.8051$ &$0.736$ & $0.732$ & $0.338$ & $0.761$ & $0.774$ \\
		MTurk-771  & $0.650$ & $0.6712$ &$0.593$ & $0.623$ & $0.199$ & $0.665$ & $0.671$ \\
		WS-353-SIM & $0.682$ & $0.7883$ &$0.713$ & $0.702$ & $0.220$ & $0.720$ & $0.720$\\
		MC-30      & $0.749$ & $0.8112$ &$0.799$ & $0.726$ & $0.330$ & $0.735$ & $0.776$ \\
		MTurk-287  & $0.649$ & $0.6645$ &$0.591$ & $0.631$ & $0.295$ & $0.674$ & $0.634$ \\
		\hline
		Average & $0.552$ & $0.6141$ & $0.519$ & $0.538$ & $0.207$ & $0.570$ & $0.579$ \\
		\hline
	\end{tabular}
	}
\end{table}

Word analogy test is introduced in \cite{mikolov13word2vec_b} and looks for the answers of the questions that are in the form of ``X is to Y, what Z is to ?" by applying simple arithmetic operations to vectors of words X, Y and Z. We present precision scores for the word analogy tests in Table~\ref{table:scores}. It can be seen that the alternative approaches that aim to improve interpretability, have poor performance on the word analogy tests. However, our proposed method has comparable performance with the original GloVe embeddings. Our method outperforms GloVe in semantic analogy test set and in overall results, while GloVe performs slightly better in syntactic test set. This comparable performance is mainly due to the cost function of our proposed method that includes the original objective of the GloVe.

To investigate the effect of the additional cost term on the performance improvement in the semantic analogy test, we present Table~\ref{table:scores2}. In particular, we present results for the cases where i) all questions in the dataset are considered, ii) only the questions that contains at least one concept word are considered, iii) only the questions that consist entirely of concept words are considered. We note specifically that for the last case, only a subset of the questions under the semantic category \verb|family.txt| ended up being included. We observe that for all three scenarios, our proposed algorithm results in an improvement in the precision scores. However, the greatest performance increase is seen for the last scenario, which underscores the extent to which the semantic features captured by embeddings can be improved with a reasonable selection of the lexical resource from which the concept word-groups were derived.

\begin{table}
	\caption{\protect\centering Precision scores for the Analogy Test}
	\label{table:scores}
	\centering
	\begin{tabular}{lcccc}
		\hline 
		Methods & \# dims & Analg. (sem) & Analg. (syn) & Total \\
		\hline
		GloVe & $300$ & $78.94$ & $64.12$ & $70.99$\\
		Word2Vec & $300$ & $81.03$ & $66.11$ & $73.03$\\
		OIWE-IPG & $300$ & $19.99$ & $23.44$ & $21.84$\\
		SOV & $3000$ & $64.09$ & $46.26$ & $54.53$\\
		SPINE & $1000$ & $17.07$ & $8.68$ & $12.57$\\
		Word2Sense & $2250$ & $12.94$ & $19.44$ & $16.51$\\
		Proposed & $300$ & $79.96$ & $63.52$ & $71.15$\\
		\hline
	\end{tabular}
\end{table}

\begin{table}
	\caption{\protect\centering Precision scores for the Semantic Analogy Test}
	\label{table:scores2}
	\centering
	\resizebox{\columnwidth}{!}{%
	\begin{tabular}{lcccc}
		\hline
		Questions Subset & \# of Questions Seen & GloVe &  Word2Vec & Proposed\\
		\hline
		All & $8783$ & $78.94$ & $81.03$& $79.96$ \vspace{0.1cm} \\ 
		At least one & \multirow{2}{*}{$1635$} & \multirow{2}{*}{$67.58$} & \multirow{2}{*}{$70.89$} & \multirow{2}{*}{$67.89$} \vspace{-0.1cm} \\
		concept word & & & \vspace{0.1cm} \\ 
		All concept words & $110$ & $77.27$ & $89.09$ & $83.64$ \\
		\hline
	\end{tabular}
	}
\end{table}

Lastly, we compare the model performances on a sentence-level binary classification task based on the Stanford Sentiment Treebank which consists of thousands of movie reviews \cite{socher13treebank} and their corresponding sentiment scores. We omit the reviews with scores between 0.4 and 0.6, resulting in 6558 training, 824 development, and 1743 test samples. We represent each review as the average of the vectors of its words. We train an SVM classifier on the training set, whose hyperparameters were tuned on the validation set. Classification accuracies on the test set are presented in Table \ref{table:sentiment}. The proposed method outperforms the original embeddings and performs on par with the SOV. Pretrained Word2Sense embeddings outperform our method, however it has the advantage of training on a larger corpus. This result along with the intrinsic evaluations show that the proposed imparting method can significantly improve interpretability without a drop in performance.

\begin{table}
	\caption{\protect\centering Accuracies (\%) for Sentiment Classification Task}
	\label{table:sentiment}
	\centering
\resizebox{\columnwidth}{!}{%
	\begin{tabular}{ccccccc}
		\hline
		GloVe & Word2Vec & OIWE-IPG & SOV & SPINE & Word2Sense & Proposed \\
		$72.62$ & $ 77.91 $ & $73.47$ & $77.45$ & $74.07$ & $81.32$  & $78.31$\\
		\hline
	\end{tabular}
	}
\end{table}

In addition to the comparisons above, we also compare our method against two projection based simple baselines. First, we project the GloVe vectors onto randomly selected 300 tokens which results in a new 300 dimensional embedding space ($RTP$ - Random Token Projections). We repeat this process for 10 times independently and report the average results. Second, we calculate the average of the vectors for the words in each of the 300 word-groups extracted from Roget's Thesaurus. Then, we project the original embeddings onto these average vectors to obtain $RCP$ (Roget Center Projections). Table \ref{table:baselines} presents the results for the task performance and interpretability evaluations for these two baselines along with the original GloVe embeddings and imparted embeddings. Although, these simple projection based methods are able to improve interpretability, they distort the inner structure of the embedding space and reduce its performance significantly.

\begin{table}
    \centering
	\caption{\protect\centering Comparisons against Random Token/Roget Center Projection baselines}
	\label{table:baselines}	
	\begin{tabular}{lcccc}
		\hline
		Task & GloVe & RTP & RCP & Proposed \\ \hline 
	    Sem. Analogy & $78.94$ & $2.84$ & $12.35$ & $79.96$ \\
	    Syn. Analogy & $64.12$ & $4.42$ & $19.70$ & $63.52$ \\ 
	    Word Sim. & $0.552$ & $0.115$ & $0.252$ & $0.579$  \\ 
	    Sentiment Clf. & $72.62$ & $51.76$ & $ 77.28$ & $78.31$ \\\hline 
	    Interp.$_{nim=5}$ & $29.23$ & $47.62$ & $64.95$ & $65.19$ \\  
	    Interp.$_{nim=10}$ & $19.22$ & $33.83$ & $54.02$ & $45.87$  \\ \hline 
	\end{tabular}
\end{table}

\subsection{Effect of Weighting Parameter \textit{k}} \label{k_selection}

\begin{figure*}[!t]
	\includegraphics[width=\textwidth]{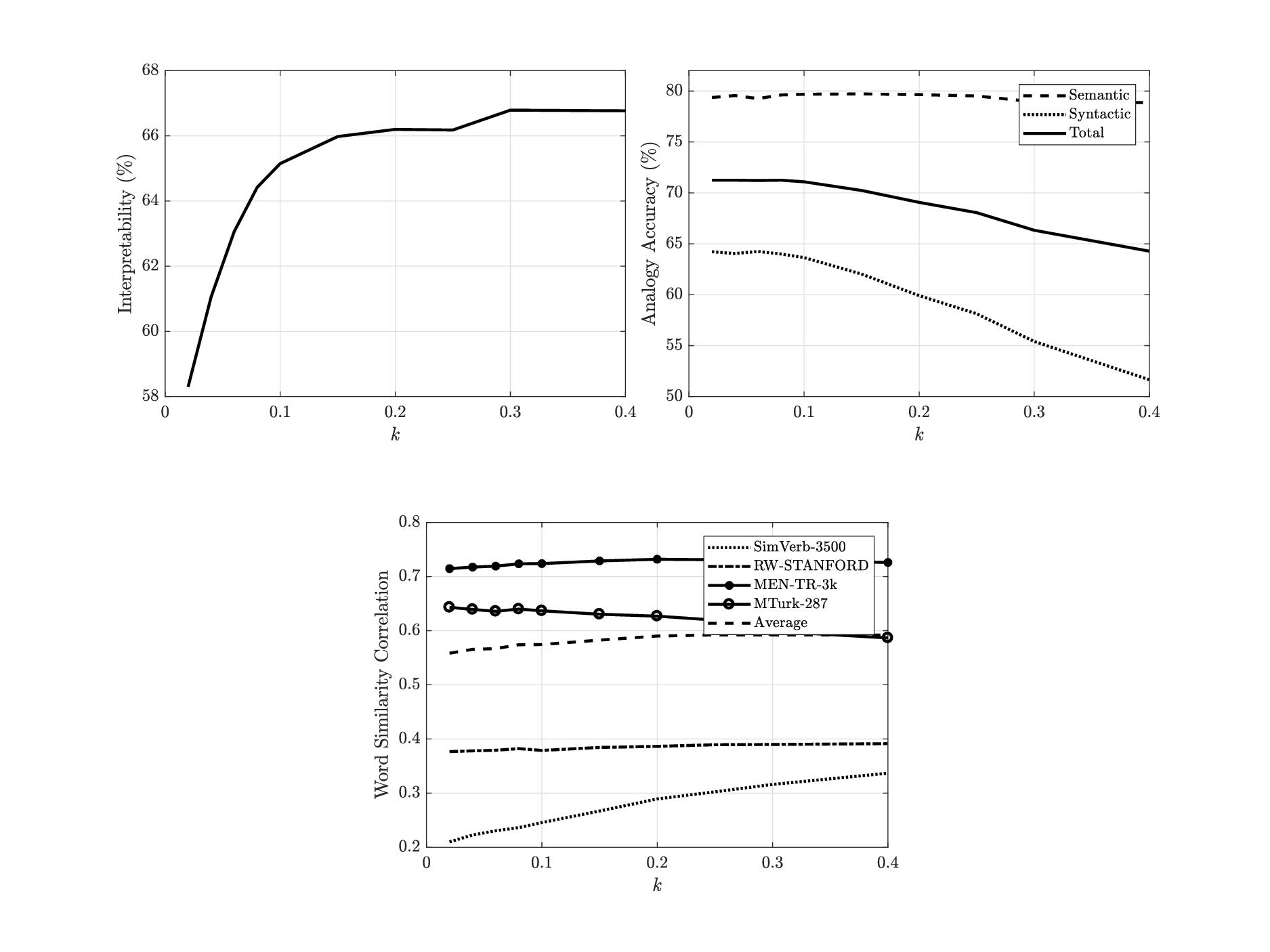}	
	\caption{Effect of the weighting parameter $k$ is tested using interpretability (top left, $n_{min}$=5, $\lambda$=5), word analogy (top right) and word similarity (bottom) tests for $k \in [0.02-0.4]$.}
	\label{fig:k_values}
\end{figure*}

The results presented in the above subsections are obtained by setting the model weighting parameter $k$ to $0.1$. However, we have also experimented with different $k$ values to find the optimal value for the evaluation tests and to determine the effects of our model parameter $k$ to the performance. Fig.~\ref{fig:k_values} presents the results of these tests for $k\in [0.02-0.4]$ range. Since parameter $k$ adjusts the magnitude of the influence for the concept words (ie. our additional term), average interpretability of the embeddings increases when $k$ is increased. However, the increase in the interpretability saturates and we almost hit the diminishing returns beyond $k = 0.1$. It can also be observed that by further increasing $k$ beyond $0.3$ no additional increase in the interpretability can be obtained. This is because interpretability measurements are based on the ranking of words in the embedding dimensions. With increasing $k$, concept words (from Roget's Thesarus) are strongly forced to have larger values in their corresponding dimensions. However, their ranks will not further increase significantly after they all reach to the top. In other words, a value of $k$ between $0.1$ and $0.3$ is sufficient in terms of interpretability. 

We now move to test whether high $k$ values will harm the underlying semantic structure or not. To test this, our standard analogy and word similarity tests that are given in the previous subsections are deployed for a range of $k$ values. Analogy test results show that larger $k$ values reduce the performance of the resulting embeddings on syntactic analogy tests, while semantic analogy performance is not significantly affected. For the word similarity evaluations, we have used 13 different datasets, however, in Fig.~\ref{fig:k_values} we present four of them as representatives along with the average of all 13 test sets to simplify the plot. Word similarity performance slightly increases for most of the datasets with increasing $k$, while performance slightly decreases or does not change for the others. On average, word similarity performance increases slowly with increasing $k$ and it is less sensitive to the variation of $k$ value than the interpretability and analogy tests. 

Combining all these experiments and observations, empirically setting $k$ to $0.1$ is a reasonable choice to compromise the trade-off since it significantly improves interpretability without sacrificing analogy/similarity performances. 

\subsection{Effect of Number of Dimensions} \label{dim_selection}

All the results presented above for our imparting method are for 300 dimensional vectors which is a common choice to train word embeddings. We trained the imparting method with the 300 word-groups extracted from Roget's Thesaurus for all experiments in order to make full use of embedding dimensions for interpretability. To investigate the effect of number of dimensions on interpretability and performance of the imparted word embeddings, we also trained the proposed method using 200 and 400 dimensions. In both cases, we make full use of the embedding dimensions. To achieve this, we extracted 200 and 400 word-groups from Roget's Thesaurus by discarding categories that have less than 76 and 36 words respectively.

\begin{table}
    \centering
	\caption{\protect\centering Effect of embedding dimension to the imparting performance.}
	\label{table:dim_count}
	\resizebox{\columnwidth}{!}{
	\begin{tabular}{lcccccc}
		\hline 
		 & \multicolumn{2}{c}{200 dim.} & \multicolumn{2}{c}{300 dim.} & \multicolumn{2}{c}{400 dim.} \\ \hline
		Task & original & proposed & original & proposed & original & proposed \\ \hline 
	    Sem. Analogy & $77.13$ & $77.64$ & $78.94$ & $79.96$ & $80.46$ & $80.95$ \\
	    Syn. Analogy & $61.91$ & $61.29$ & $64.12$ & $63.52$ & $64.70$ & $64.06$ \\ 
	    Word Sim. & $0.541$ & $0.557$ & $0.552$ & $0.579$ & $0.554$ & $0.586$ \\ 
	    Sentiment Clf. & $77.51$ & $77.51$ & $77.34$ & $78.26$ & $77.28$ & $77.40$ \\ \hline 
	    Interp.$_{nim=5}$ & $29.64$ & $69.91$ & $29.23$ & $65.19$ & $26.85$ & $61.81$ \\  
	    Interp.$_{nim=10}$ & $19.73$ & $51.09$ & $19.22$ & $45.87$ & $17.29$ & $40.42$ \\ \hline 
	\end{tabular}
	}
\end{table}

Table \ref{table:dim_count} presents the results for interpretability and performance evaluations of the GloVe and imparted embeddings for 200, 300 and 400 dimensions. For the word similarity evaluations, results are averaged across the 13 different datasets. On the performance evaluations, imparted embeddings perform on par with the original embeddings regardless of the dimensionality. It can be seen that performance on the intrinsic tests slightly improve with increasing dimensionality for both embeddings while the performance on the classification task does not change significantly. For the interpretability evaluations, the trend is the opposite. In general, interpretability decreases with increasing dimensionality since it is more difficult to consistently achieve good interpretability in more dimensions. However, imparted embeddings are significantly more interpretable than the original embeddings in all cases. Based on these results, we argue that 300 is a decent selection for dimensionality in terms of performance, interpretability and computational efficiency.

\subsection{Design of Function g(x)} \label{g_selection}

As presented in Section~\ref{sec:infl}, our proposed method encourages the trained word vectors to have larger values if the underlying word is semantically close to a collection of concept word-groups, one assigned to each vector dimension. However, a mechanism is needed to control amounts of these inflated values. To facilitate and control this behavior, a function $g(x)$ should be used. This function serves to increase the total cost incurred if the value of the $l$th dimension for the two vector representations $\vec{w}_{i,l}$ and $\vec{\tilde{w}}_{j,l}$ for a concept word $\vec{w}_i$ with $i \in F_l$ fails to be large enough. In this subsection, we will elaborate more on the design and selection methodology of this function $g(x)$.

First of all, it is very obvious to see that $g(x)$ should be a positive monotone decreasing function because the concept words taking small values in the dimension that corresponds to word groups that they belong to, should be penalized more harshly so that they are forced to take larger values. On the other hand, if their corresponding dimensions are large enough, contribution to the overall cost term coming from this objective should not increase much further. In other words, value of $g(x)$ should go to positive infinity as x decreases and go to 0 as x increases. Outstanding natural candidates for $g(x)$ with these features are then exponential decay and reciprocal of polynomials with odd degrees such as $1/x$ or $1/(x^3+1)$. Although these polynomials satisfy the requirement that the function be decaying, for negative values of x they give negative values which undesirable causes the cost to decrease. Therefore, polynomials can only be used for the positive x values. What is left is exponential decays. In exponential case, there is no concern for $g(x)$ taking negative values. All we need to do is just to adjust the decaying rate. Too fast decays can make the additional objective lose its meaning while too slow decays can break the structure of the general objective function by disproportionately putting more emphasis on our proposed modification term that is added to the original cost term of the embedding. (Here, it should also be noted that the adjustment of this parameter needs to be done in conjunction with parameter $k$ that controls the blending of the two sub-objectives, which has also been analyzed in detail in Section~\ref{k_selection}.)

To further study several alternatives, we have also considered piece-wise functions composed of both decaying exponentials and polynomials to facilitate and study properties of both. After several experiments for hyperparameter estimation, we conclude that the functions like $(1/2)\exp{(-2x)}$ or $(1/3)\exp{(-3x)}$ give best results and we have proposed the function in Eq.~\eqref{eq:g} which switches continuously from a decaying exponential to a reciprocal of a polynomial when $x$ becomes greater than $0.5$ (transition from $(1/2)\exp{(-2x)}$ to $1/(4ex)$ is adjusted such that $g(x)$ is continuous). A piece-wise function is formed such that its polynomial part decays more slowly than its exponential part does so that the objective function can keep pushing the words with small $x$ values a little bit longer. Furthermore, $\exp{(-x)}$ has also been tried but since its decay is too slow and it takes large values, GloVe did not converge properly. On the other hand, faster decays also did not work since they quickly neutralize the additional interpretability cost term.

\begin{figure}[!h]
    \centering
	\includegraphics[width=\columnwidth]{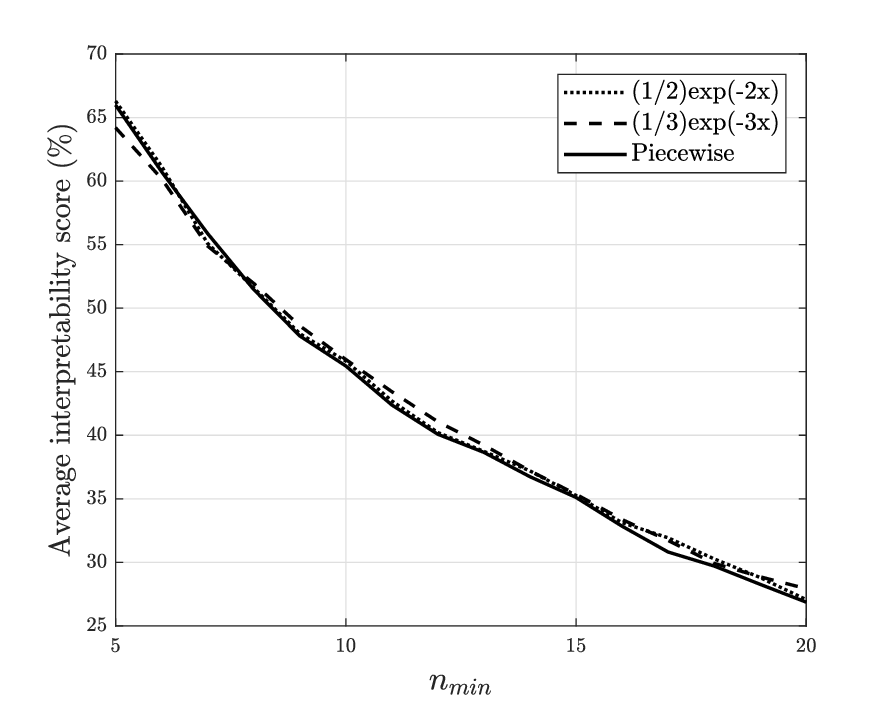}
	\newline\newline
	\caption{Interpretability scores averaged over 300 dimensions for the proposed method for different forms of function $g(x)$.}
	\label{fig:g_forms}
\end{figure}

Experimental results for interpretability scores averaged over 300 dimensions are presented in Fig.~\ref{fig:g_forms} for $g(x)$ options $(1/2)\exp{(-2x)}$, $(1/3)\exp{(-3x)}$, and the piece-wise function given in Eq.~\eqref{eq:g} ($k$ is kept at $0.10$). Experiments show that there is not much difference between single exponential function and piece-wise function but the single exponential decay and piece-wise $g(x)$ with decay rates of $-2$ seems to be slightly better. To sum up, one can choose the piece-wise option since it includes both natural options.

\section{Conclusion}
\label{sec:conclusion}

We presented a novel approach to impart interpretability into word embeddings. We achieved this by encouraging different dimensions of the vector representation to align with predefined concepts, through the addition of an additional cost term in the optimization objective of the GloVe algorithm that favors a selective increase for a pre-specified input of concept words along each dimension.

We demonstrated the efficacy of this approach by applying qualitative and quantitative evaluations based on both automated metrics and on manual human annotations for interpretability. We also showed via standard word-analogy and word-similarity tests that the semantic coherence of the original vector space is preserved, even slightly improved. We have also performed and reported quantitative comparisons with several other methods for both interpretability increase and preservation of semantic coherence. Upon inspection of Fig.~\ref{fig:interpretability} and Tables \ref{table:similarity_scores}, \ref{table:scores}, and \ref{table:scores2} altogether, it should be noted that our proposed method achieves both of the objectives simultaneously, increased interpretability and preservation of the intrinsic semantic structure.

An important point was that, while it is expected for words that are already included in the concept word-groups to be aligned together since their dimensions are directly updated with the proposed cost term, it was also observed that words not in these groups also aligned in a meaningful manner without any direct modification to their cost function. This indicates that the cost term we added works productively with the original cost function of GloVe to handle words that are not included in the original concept word-groups, but are semantically related to those word-groups.
The underlying mechanism can be explained as follows. While the outside lexical resource we introduce contains a relatively small number of words compared to the total number of words, these words and the categories they represent have been carefully chosen and in a sense, ``densely span" all the words in the language. By saying ``span", we mean they cover most of the concepts and ideas in the language without leaving too many uncovered areas. With ``densely" we mean all areas are covered with sufficient strength. In other words, this subset of words is able to constitute a sufficiently strong skeleton, or scaffold. Now remember that GloVe works to align or bring closer related groups of words, which will include words from the lexical source. So the joint action of aligning the words with the predefined categories (introduced by us) and aligning related words (handled by GloVe) allows words not in the lexical groups to also be aligned meaningfully. We may say that the non-included words are ``pulled along" with the included words by virtue of the ``strings" or ``glue" that is provided by GloVe. In numbers, the desired effect is achieved by manipulating less than only 0.05\% of parameters of the entire word vectors. Thus, while there is a degree of supervision coming from the external lexical resource, the rest of the vocabulary is also aligned indirectly in an unsupervised way. This may be the reason why, unlike earlier proposed approaches, our method is able to achieve increasing interpretability without destroying underlying semantic structure, and consequently without sacrificing performance in benchmark tests.

Upon inspecting the 2nd column of Table~\ref{table:effect_on_top_words_bad}, where qualitative results for concept \verb|TASTE| are presented, another insight regarding the learning mechanism of our proposed approach can be made. Here it seems understandable that our proposed approach, along with GloVe, brought together the words \verb|taste| and \verb|polish|, and then the words \verb|Polish| and, for instance, \verb|Warsaw| are brought together by GloVe. These examples are interesting in that they shed insight into how GloVe works and the limitations posed by polysemy. It should be underlined that the present approach is not totally incapable of handling polysemy, but cannot do so perfectly. Since related words are being clustered, sufficiently well-connected words that do not meaningfully belong along with others will be appropriately ``pulled away" from that group by several words,  against the less effective, inappropriate pull of a particular word. Even though \verb|polish| with lowercase ``p" belongs where it is, it is attracting \verb|Warsaw| to itself through polysemy and this is not meaningful. Perhaps because \verb|Warsaw| is not a sufficiently well-connected word, it ends being dragged along, although words with greater connectedness to a concept group might have better resisted such inappropriate attractions. Addressing polysemy and meaning conflation deficiency is beyond the scope and intention of our proposed model that represents all senses of a word with a single vector. However, our model may open up new directions of research in the intermingled studies on representing word senses and contextualized word embeddings~ \cite{jang18contextinterpretable,camachocollados18surveywordrep}. For example, the associated concepts of our model assigned to the word embedding dimensions may be chosen such that different senses are assigned to different dimensions.

In this study, we used the GloVe algorithm as the underlying dense word embedding scheme to demonstrate our approach. However, we stress that it is possible for our approach to be extended to other word embedding algorithms which have a learning routine consisting of iterations over cooccurrence records, by making suitable adjustments in the objective function. Since word2vec model is also based on the coocurrences of words in a sliding window through a large corpus, we expect that our approach can also be applied to word2vec after making suitable adjustments, which can be considered as an immediate future work for our approach. Although the semantic concepts are encoded in only one direction (positive) within the embedding dimensions, it might be beneficial to pursue future work that also encodes opposite concepts, such as good and bad, in two opposite directions of the same dimension.

The proposed methodology can also be helpful in computational cross-lingual studies, where the similarities are explored across the vector spaces of different languages~\cite{mikolov13crossLingual,senel17crossLingual}.

\section*{Acknowledgements}
We thank Dr. Tolga Cukur (Bilkent University) for fruitful discussions. We would also like to thank the anonymous reviewers for their many comments which significantly improved the quality of our manuscript. We also thank the anonymous human subjects for their help in undertaking the word intrusion tests. H. M. Ozaktas acknowledges partial support of the Turkish Academy of Sciences.

\ifCLASSOPTIONcaptionsoff
  \newpage
\fi



%

\end{document}